\documentclass[runningheads]{llncs}
\usepackage{graphicx}
\usepackage[utf8]{inputenc}
\usepackage{amssymb}
\usepackage{amsmath}
\usepackage{booktabs}
\usepackage{tabularx}
\usepackage{tikz}

\usepackage{pgfplots}
\usepackage{pgf-umlsd}
\usepackage{ifthen}
\usepackage{subcaption}
\usetikzlibrary{3d} 
\usepackage{multicol}

\begin{document}
\title{Lane Detection and Classification using Cascaded CNNs\thanks{The GPU used has been donated by the NVIDIA corporation.}}
%
%
\author{Fabio Pizzati\inst{1}\and
Marco Allodi\inst{2}\and
Alejandro Barrera\inst{3}\and
Fernando García\inst{3}}
\authorrunning{F. Pizzati et al.}

\institute{University of Bologna,
 Viale Risorgimento 2, 40136 Bologna BO, Italy\\
\email{fabio.pizzati2@unibo.it}\\ \and
University of Parma, Via delle Scienze, 181 / a, 43124 Parma PR, Italy \\
\email{marco.allodi1@studenti.unipr.it}
\and
Universidad Carlos III de Madrid, Av. de la Universidad, 30, 28911 Leganés, Madrid, Spain\\
\email{alebarre@pa.uc3m.es, fegarcia@ing.uc3m.es}}
\maketitle              
\begin{abstract}
Lane detection is extremely important for autonomous vehicles. For this reason, many approaches use lane boundary information to locate the vehicle inside the street, or to integrate GPS-based localization. As many other computer vision based tasks, convolutional neural networks (CNNs) represent the state-of-the-art technology to indentify lane boundaries. However, the position of the lane boundaries w.r.t. the vehicle may not suffice for a reliable positioning, as for path planning or localization information regarding lane types may also be needed. In this work, we present an end-to-end system for lane boundary identification, clustering and classification, based on two cascaded neural networks, that runs in real-time. To build the system, 14336 lane boundaries instances of the TuSimple dataset for lane detection have been labelled using 8 different classes. Our dataset and the code for inference are available online.

\end{abstract}
\section{Introduction}\label{introduction}

In autonomous driving, a deep understanding of the surrounding environment is vital to safely drive the vehicle. For this reason, a precise interpretation of the visual signals is necessary to identify all the components essential for navigation. One of them of particular importance is lane boundary position, which is needed to avoid collisions with other vehicles, and to localize the vehicle inside the street. Besides easing localization, lane detection is employed in many ADAS for lane departure warning and lane keeping assist. As many others computer vision based tasks, lanes boundaries detection accuracy has been significantly improved after the introduction of deep learning. The majority of recent systems, indeed, use convolutional neural networks to process sensorial data and infer high-level information relative to lanes. Some of them process LiDAR data to exploit differences in lane markings reflectivity \cite{caltagirone2017fast,bai2018deep}. However, LiDARs are extremely expensive, thus not always available on a vehicle. On the other hand, cameras are cheaper, and they make it possible to exploit chromatic differences on the road surface. Among the deep learning based lane detection approaches that rely solely on visual data, there is significant interest in lane marking detection \cite{DBLP:journals/corr/abs-1809-03994,tian2018lane,li2017deep,Lee2017VPGNetVP,zang2018traffic}. In \cite{tian2018lane}, a modified version of Faster R-CNN \cite{ren2015faster} is used to identify road patches that belong to lane markings. Many of those patches are then joined to obtain a complete representation of the marking. However, the system runs at approximately 4 frames per seconds, so it is not suitable for real-time elaboration, that is often a hard requirement for high-speed driving. In other works \cite{Lee2017VPGNetVP,zang2018traffic,John2018} lane markings detection and classification is achieved using fully-convolutional networks \cite{long2015fully}, and this enables more complex path planning tasks, where lane changes could be considered. Nonetheless, a comprehensive understanding of lane boundaries may be needed for path planning, so it may be necessary to join the detected markings with post processing algorithms. An alternative approach is to directly identify the boundaries, in order to reduce post-processing times. In \cite{neven2018towards,pan2018spatial,zhang2018geometric,kim2017end}, fully-convolutional networks are used to obtain a pixelwise representation of lane boundaries. A slightly different approach is proposed in \cite{Chougule}, where a CNN is used to estimate polylines points, in order to solve fragmentation issues that often occurr in segmentation networks. They classify the obtained boundaries, but only in terms of position w.r.t. the ego vehicle, so no information regarding the lane boundary type (e.g. dashed, continuous) is extracted. Similarly to \cite{Chougule}, Ghafoorian et al. \cite{ghafoorian2018gan} exploit adversarial training to reduce fragmentation. In \cite{wvangansbeke_2019}, an end-to-end approach is proposed, where lane boundary parameters are directly estimated by a CNN.
As lane boundaries position, also lane boundaries types could be exploited to achieve a high grade of scene understanding. For example, knowing if a lane is dashed is indispensable for a lane change. Nonetheless, there is little interest in literature for simultaneous lane boundary identification and classification using deep learning. This could be caused by the lack of datasets that contains both information. For this reason, we extended a lane detection dataset with lane class annotations. Then, we developed a novel approach, based on the concatenation of multiple neural networks, that is used to perform lane boundary instance segmentation and classification, in an end-to-end deep learning based fashion. Our system satisfy real-time constraints on a NVIDIA Titan Xp GPU. Code for inference and pretrained models are available online.

\section{Method}
\begin{figure}[h]
\hfill
\centering
\includegraphics[height=64px]{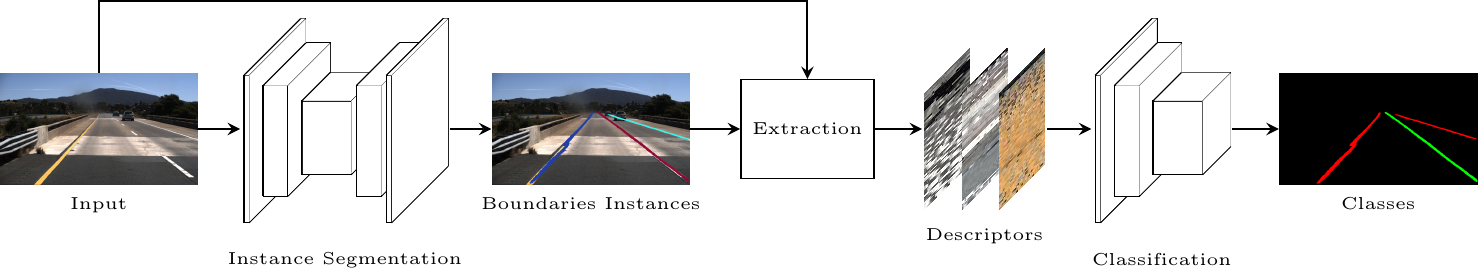}
\hfill
\caption{System overview}\label{figure:system}
\end{figure}

Our method is composed by two main sections, as presented in figure \ref{figure:system}. As a first step, we train a CNN for lane boundary instance segmentation. Then, we extract a descriptor for each detected lane boundary and process it with a second CNN.
\subsection{Instance Segmentation}

As discussed in section \ref{introduction}, several state-of-the-art approaches employ pixelwise classifications in order to differentiate pixels belonging to lane boundaries and background. In our case, different approaches are possible, so several design guidelines have been defined. First of all, we train the CNN to recognize lane boundaries, rather than lane markings. Doing this, it is indeed avoidable to group different lane markings in a lane boundary, considerably saving processing times. For similar reasons, we perform instance segmentation on lane boundaries instead of semantic segmentation. In this way, it is possible to distinguish different lane boundaries without relying on clustering algorithms. The state-of-the-art network for instance segmentation is Mask R-CNN \cite{he2017mask}. However, two-step networks like Mask R-CNN are tipically not amenable to use in real-time application, as they are typically slower than single-step ones. Furthermore, they are usually employed to detect objects easily enclosable in rectangular boxes. Being lane boundaries appeareance heavily influcenced by the perspective effects, other kinds of architectures should be preferred. Taking into account the previous assumptions, a fully-convolutional network has been trained.

We choose ERFNet \cite{romera2018erfnet} as our baseline model, as it is the model that, at the time of writing, has the best performances among real-time networks in the Cityscapes Dataset benchmark for semantic segmentation. As in all deep-learning based approaches, large amounts of images and annotations are crucial to achieve correct predictions and to avoid overfitting. For this reason, the TuSimple dataset for lane detection has been used. It is composed by 6408 1280\(\times\)720 images, divided in 3626 for training, and 2782 for testing. 410 images extracted from the training set have been used as validation set during training. The main peculiarity in the TuSimple dataset is that entire lane boundaries are annotated, rather than lane markings. This makes the TuSimple dataset ideal for our needs. The lane boundaries are represented as polylines. 
In order to avoid clustering via post-processing, it is possible to make use of the loss function presented in \cite{hsu2018}, that is based on the Kullback-Leibler divergence minimizaton between the output probability distributions associated to pixels belonging to the same lane boundary instance. Please note that we do not address an unlimited number of possible instances for lane boundaries, as we decided to detect only the ego-lane boundaries, and the ones of the lanes on the sides of the ego-lane. Considering that two boundaries are shared for different lanes, we set a fixed maximum number of detected boundaries to 4. However, directly training the network with \cite{hsu2018} ultimately leads to gradient explosion and loss divergence. For this reason, the curriculum learning strategy \cite{bengio2009curriculum} has been used to achieve convergence. In fact, in a first step a binary cross entropy loss has been used to train the network to distinguish between points belonging to a generic lane boundary and background. The resulting model is fine-tuned using \cite{hsu2018} as loss function. 
The network has been trained using the images in the dataset resized at \(512\times256\) resolution, for 150 epochs. This resolution leaded to satisfying results, while keeping the computational cost low. In order to represent the ground truth data as images, the polylines in the dataset have been projected with a fixed width of 5px on semantic maps of size \(512\times256\). We use the Adam optimizer, with learning rate \(5\cdot10^{-4}\), and polynomial learning rate decay with exponent set to 0.9.



\subsection{Classification}
To the best of our knowledge, there are currently no publicly available datasets where entire lane boundaries with class-related information are annotated. For this reason, all the lanes in the TuSimple dataset have been manually classified using 8 different classes: \textit{single white continuous}, \textit{double white continuous}, \textit{single yellow continuous}, \textit{double yellow continuous}, \textit{dashed}, \textit{double-dashed}, \textit{Botts' dots} and \textit{unknown}. The obtained annotations are available online at \texttt{https://github.com/fabvio/TuSimple-lane-classes}.

\begin{figure}[h]
\hfill
\centering
\begin{subfigure}{.33\linewidth}
	\centering

  \includegraphics[height=31px]{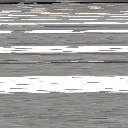}
  \caption{128\(\times\)128}
  \label{fig:sub1}
\end{subfigure}%
\hfill
\begin{subfigure}{.33\linewidth}
\centering

  \includegraphics[height=31px]{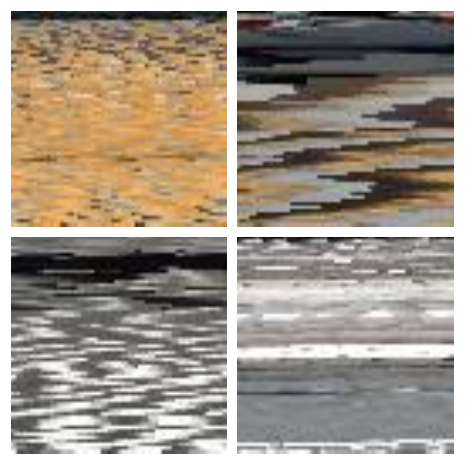}
  \caption{64\(\times\)64}
  \label{fig:sub2}
\end{subfigure}
\hfill
\begin{subfigure}{.33\linewidth}
\centering

  \includegraphics[height=31px]{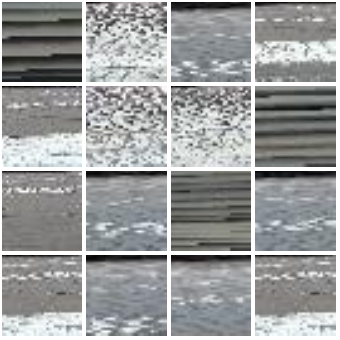}
  \caption{32\(\times\)32}
  \label{fig:sub2}
\end{subfigure}
\hfill

\caption{Descriptors of different sizes}
\label{fig:descriptors}

\end{figure}

Associating a class to each lane boundary detected could be addressed in several different ways. One possibility that has been considered in an early stage of the development was branching the instance segmentation network, to perform a pixel-wise classification of lane boundaries with dedicated convolutional layers, then fuse the outputs of the two branches. This approach has been discharged as it is memory intensive, because it requires two decoders in the same network, and it may generate inconsistencies between the detection of the two branches, that should be solved using post-processing algorithms. For example, there could be pixels classified as background from the instance segmentation branch, but classified as lane from the other branch. For this reason, we perform a classification for each lane boundary with another CNN, associating the detected boundaries to the ground truth. A problem with this approach is that each lane boundary is constituted by a different number of points in the input image. For this reason, it is difficult to extract a representation of them that is position-indipendent w.r.t. the ego vehicle. This may be essential to achieve a correct classification. Thus, we extract a descriptor for each boundary, sampling a fixed number of points from the input image which belong to the detected lane boundary. The points extracted in this way are then ordered following their index in the original image, and arranged in squared images, that are processed by the second neural network. In this way, a spatially normalized compact representation of lane boundaries is obtained, while preserving information given by visual clues such as lane markings. Furthermore, using this approach we are able to perform lane boundary instance segmentation and classification with only two inferences, in an end-to-end fashion. In fact, the descriptors of different lane boundaries detected could be grouped in batches and classified simultaneously. Examples of descriptors are shown in image \ref{fig:descriptors}.

\begin{figure}[hbt!]
\hfill
\centering
\begin{subfigure}{.7\linewidth}
	\centering

  \includegraphics[height=80px]{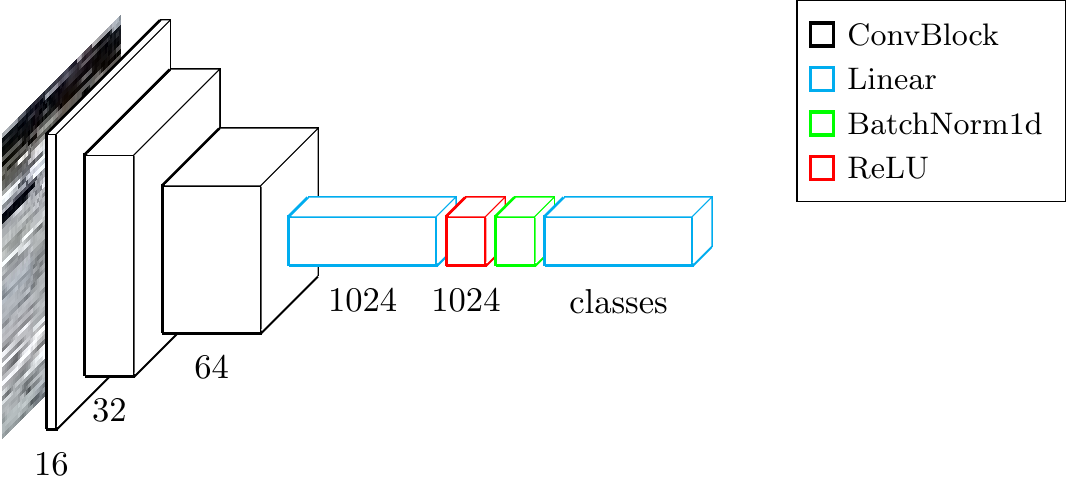}
  \caption{CNN architecture}
  \label{fig:sub1}
\end{subfigure}%
\hfill
\hfill
\begin{subfigure}{.3\linewidth}
\centering

  \includegraphics[height=80px]{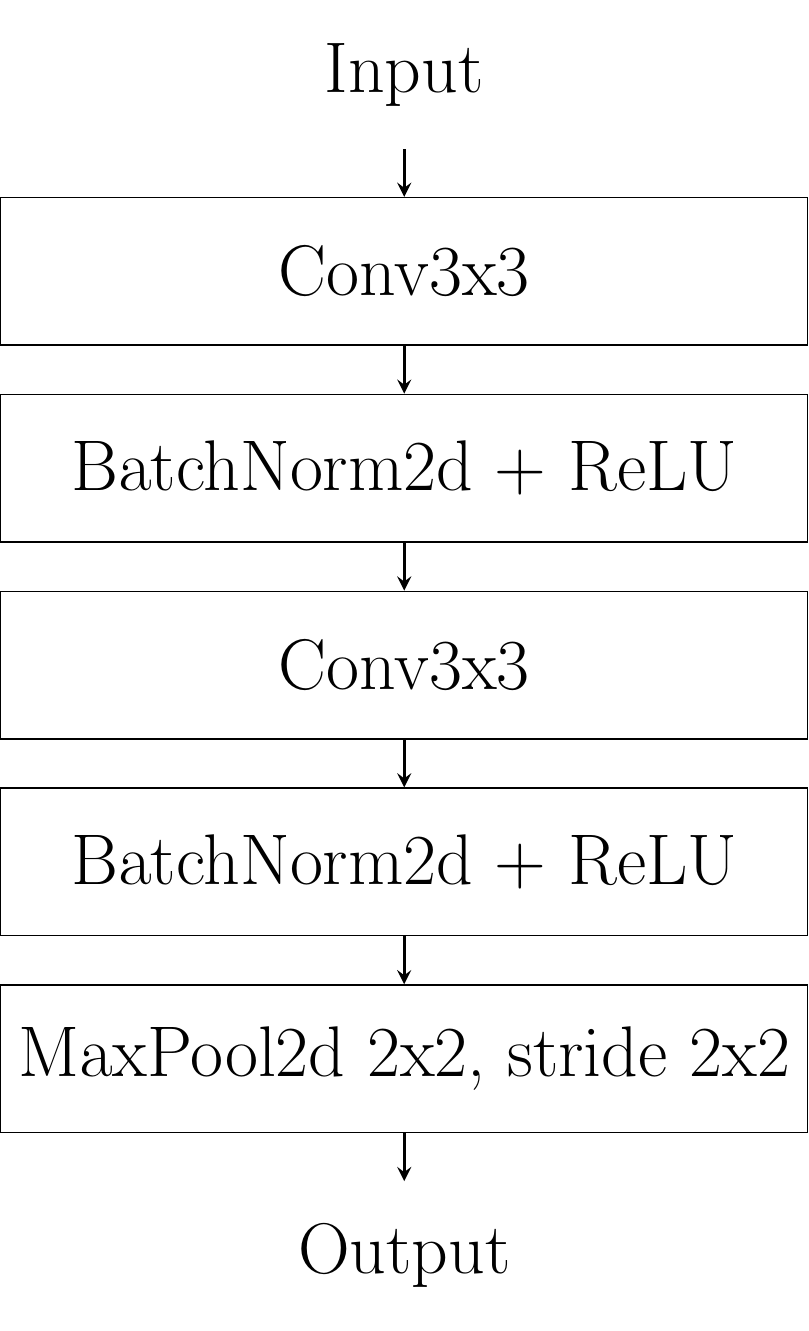}
  \caption{ConvBlock}
  \label{fig:sub2}
\end{subfigure}
\hfill

\caption{Classification Network. Output channels are listed below each layer.}
\label{fig:hnet}

\end{figure}

The architecture we use for this task is derived from H-Net \cite{neven2018towards}. A detailed description of its structure is given in image \ref{fig:hnet}. We trained this network separately from the first one. To do that, the TuSimple dataset has been processed by the instance segmentation network. Each detected lane boundary is then compared with the ground truth, and it is associated to the corresponding class if the average distance between the detected points and the ground truth is under a threshold. This is needed to filter false positives generated by the first network. In fact, only lane boundaries that are effectively in the training set have a ground truth class, while others detected by the CNN should be excluded. As a result, we obtain a set of \(\{descriptor, class\}\) objects that can be used to train the classification neural network. This has been trained with the same hyperparameters of the instance segmentation network. Examples of extracted descriptors are shown in image \ref{fig:descriptors}. Code for inference, descriptor extraction and pretrained models are publicly available at \texttt{https://github.com/fabvio/Cascade-LD}.
\section{Results}
In order to validate our method, we evaluate the performances of both networks separately. For lane boundary instance segmentation, the evaluation formula for the TuSimple benchmark is presented in equation \ref{equation:tusimple_formula}. In it, \(C_{i}\) and \(S_{i}\) are the number of correctly detected points and ground truth points in image \(i\), respectively. A point is defined as correctly detected if it has a distance w.r.t. a ground truth point under 20 pixels. Additionally, false positive and false negative lane boundaries are evaluated. Given that our detected lane boundaries have width over 1 pixel, we average the \(x\) coordinates of the detected pixels for a given row, to obtain a single value. In \ref{equation:false_positives}, \(F_{pred}\) refers to the number of erroneously detected lanes, while \(N_{pred}\) is the total number of detected lanes. In \ref{equation:false_negatives}, \(M_{pred}\) is the total number of unidentified lanes, and \(N_{gt}\) is the total number of lane boundaries annotated.

\begin{multicols}{3}
\begin{equation}\label{equation:tusimple_formula}
accuracy = \frac{\sum_{i}C_{i}}{\sum_{i}{S_{i}}}
\end{equation}\break
\begin{equation}\label{equation:false_positives}
FP = \frac{F_{pred}}{N_{pred}}
\end{equation}\break
\begin{equation}\label{equation:false_negatives}
FN = \frac{M_{pred}}{N_{gt}}
\end{equation}
\end{multicols}

We compare our instance segmentation network with the top-three approaches in the TuSimple benchmark for lane detection. We do not evaluate lanes that are composed than less of three points, in order to filter false positives. Results are presented in table \ref{table:inst_metric}. Inference times are evaluated on \(512\times256\) images. Our network is slightly less accurate than the others. However, taking into account the computational times reduction, we found this tradeoff acceptable.

\begin{table}
  \caption{\upshape{TuSimple Lane detection metrics results and comparison.}}
  \label{table:inst_metric}
  \setlength{\tabcolsep}{4\tabcolsep}
  \centering

  \begin{tabular}{ *{5}{c} }
    \toprule
    \textbf{Method} & \textbf{Accuracy} &  \textbf{FP} & \textbf{FN} & \textbf{FPS} \\
    \midrule
    Xingang Pan \cite{pan2018spatial} & \textbf{96.53} & \textbf{6.17} & \textbf{1.80} & 5.31\\
    Yen-Chang Hsu \cite{hsu2018}  & 96.50 & 8.51 & 2.69 & 55.55\\
    Davy Neven \cite{neven2018towards} & 96.40 & 23.65 & 2.76 & 52.63\\
    Ours & 95.24 & 11.97 & 6.20 & \textbf{58.93}\\
    \bottomrule
  \end{tabular}
  \vspace{1pt}
\end{table}

For classification, two different experiments are performed. In a first phase, we train the network to distinguish between two different classes: \textit{dashed} and \textit{continuous}. To do that, the \textit{single white continuous}, \textit{double white continuous}, \textit{single yellow continuous}, \textit{double yellow continuous} classes are mapped to the \textit{continuous} class. On the other hand, \textit{dashed}, \textit{Botts' dots} and \textit{double-dashed} are equally labelled as \textit{dashed}. \textit{Unknown} descriptors are ignored. In this way, it is possible to distinguish between lane boundaries that may or may not be crossed. In the second experiment, we treat the \textit{double-dashed} class as indipendent. Doing this, we could identify also the boundaries of lanes that may be crossed only in specific conditions, as highway entry or exit. An ablation study regarding the descriptor size has been performed. We evaluate classification performances on the validation set, as the test set labels for lane boundaries are not publicly available. Results are reported in table \ref{table:results}. Inference times for the classification network are around 1ms.
\begin{figure}[h!]
\begin{subfigure}{\linewidth}
  \begin{subfigure}{0.15\linewidth}
    \includegraphics[width=\textwidth]{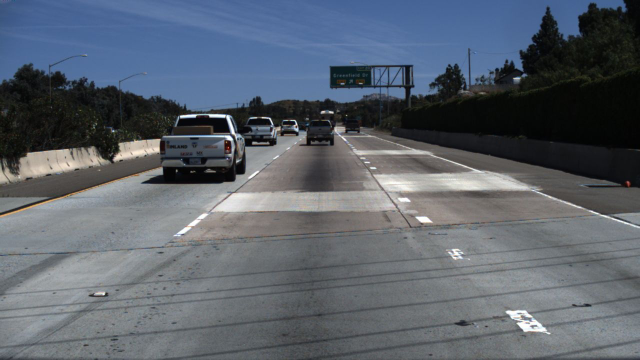}
  \end{subfigure}
  \hfill
  \begin{subfigure}{0.15\linewidth}
    \includegraphics[width=\textwidth]{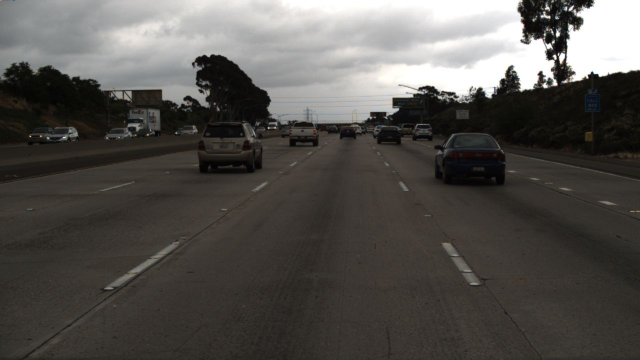}
  \end{subfigure}
  \hfill
  \begin{subfigure}{0.15\linewidth}
    \includegraphics[width=\textwidth]{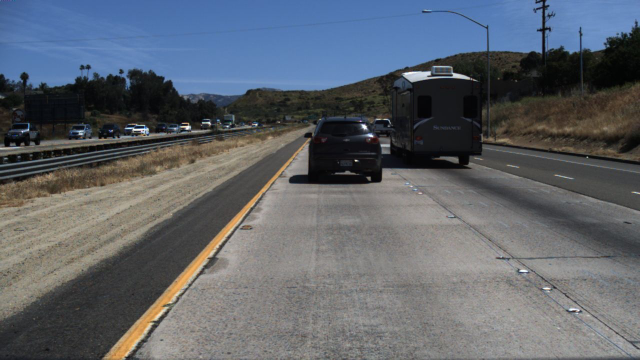}
  \end{subfigure}
  \hfill
  \begin{subfigure}{0.15\linewidth}
    \includegraphics[width=\textwidth]{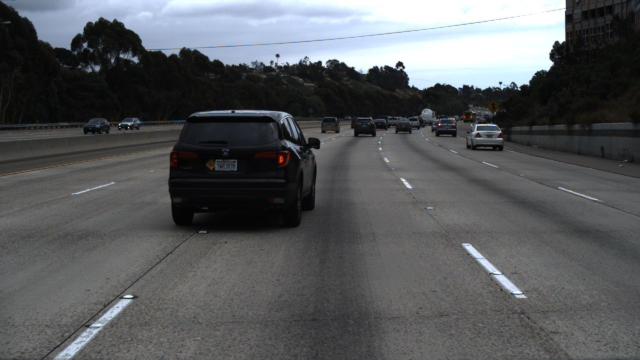}
  \end{subfigure}
  \hfill
  \begin{subfigure}{0.15\linewidth}
    \includegraphics[width=\textwidth]{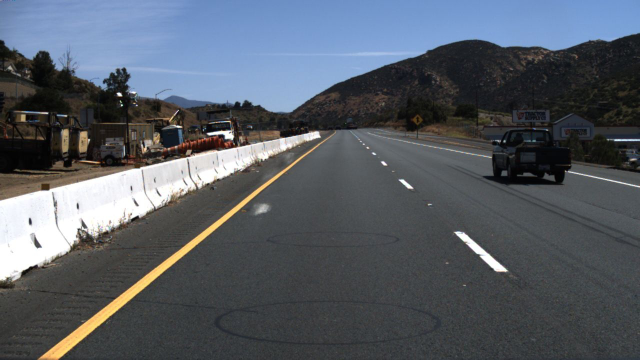}
  \end{subfigure}
  \hfill
  \begin{subfigure}{0.15\linewidth}
    \includegraphics[width=\textwidth]{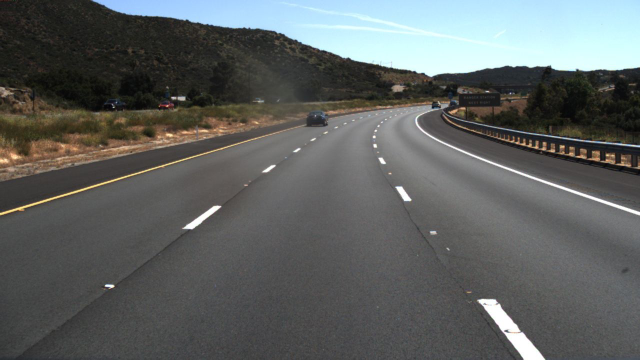}
  \end{subfigure}
  \hfill
\end{subfigure}
\begin{subfigure}{\linewidth}
  \begin{subfigure}{0.15\linewidth}
    \includegraphics[width=\textwidth]{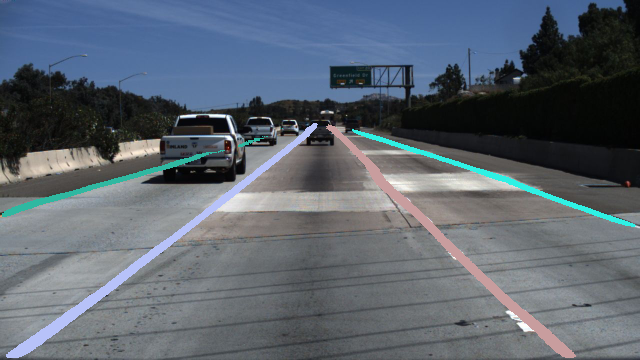}
  \end{subfigure}
  \hfill
  \begin{subfigure}{0.15\linewidth}
    \includegraphics[width=\textwidth]{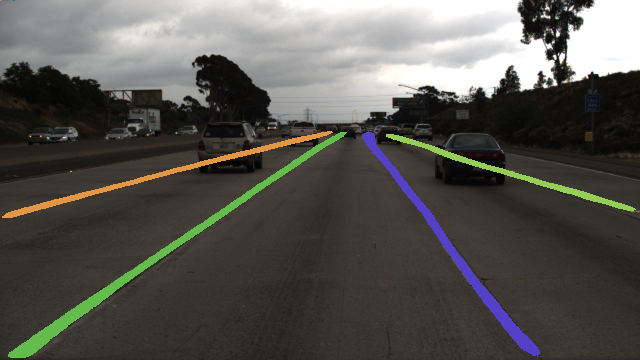}
  \end{subfigure}
  \hfill
  \begin{subfigure}{0.15\linewidth}
    \includegraphics[width=\textwidth]{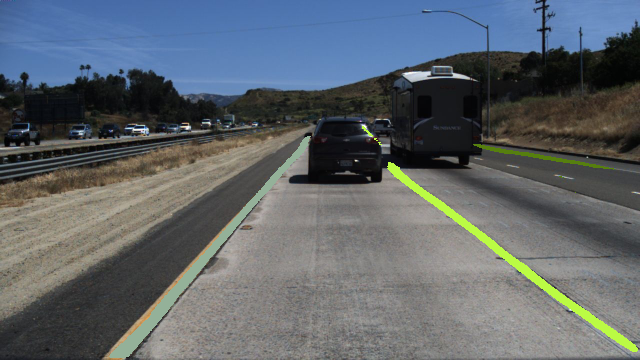}
  \end{subfigure}
  \hfill
  \begin{subfigure}{0.15\linewidth}
    \includegraphics[width=\textwidth]{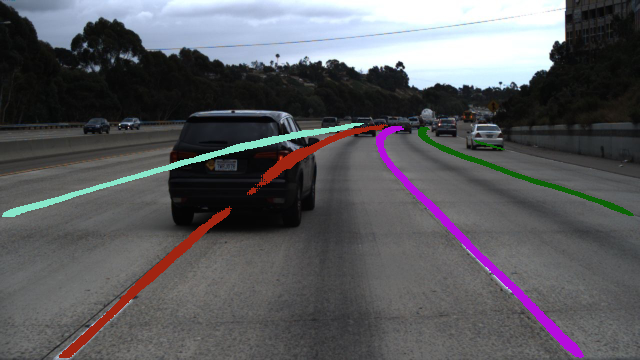}
  \end{subfigure}
  \hfill
  \begin{subfigure}{0.15\linewidth}
    \includegraphics[width=\textwidth]{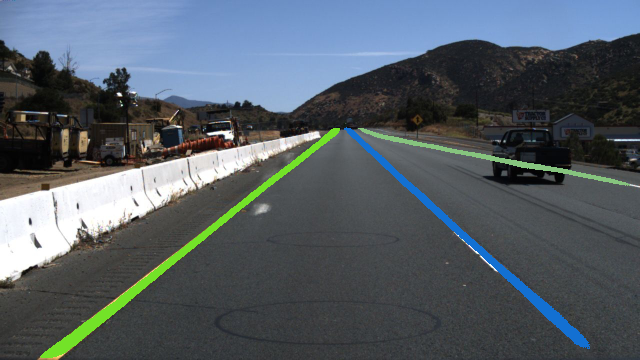}
  \end{subfigure}
  \hfill
  \begin{subfigure}{0.15\linewidth}
    \includegraphics[width=\textwidth]{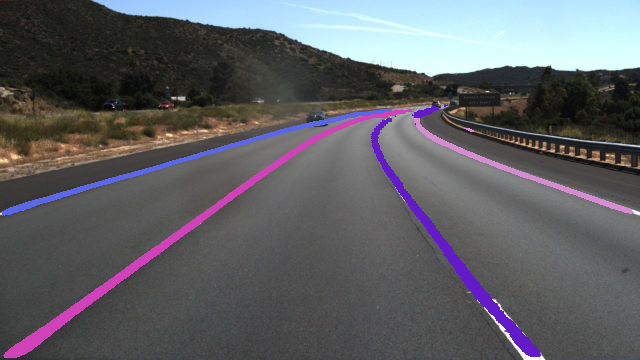}
  \end{subfigure}
  \hfill
\end{subfigure}
\begin{subfigure}{\linewidth}
  \begin{subfigure}{0.15\linewidth}
    \includegraphics[width=\textwidth]{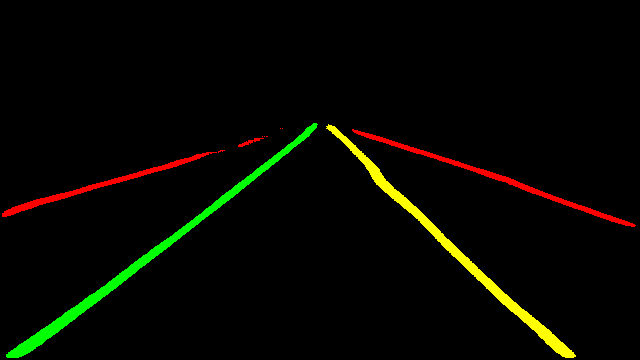}
  \end{subfigure}
  \hfill
  \begin{subfigure}{0.15\linewidth}
    \includegraphics[width=\textwidth]{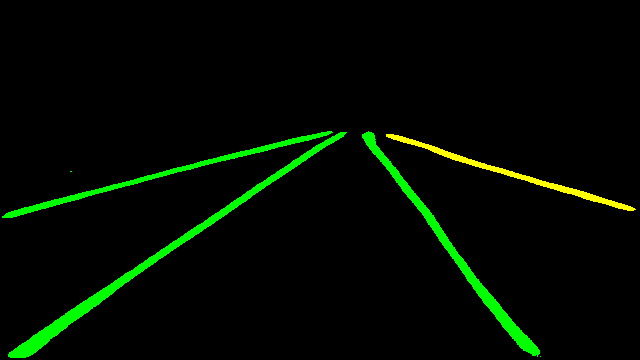}
  \end{subfigure}
  \hfill
  \begin{subfigure}{0.15\linewidth}
    \includegraphics[width=\textwidth]{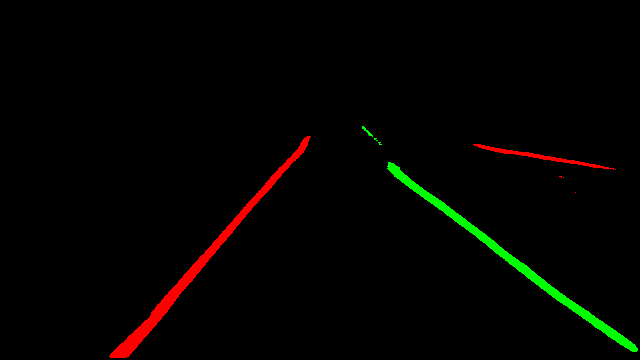}
  \end{subfigure}
  \hfill
  \begin{subfigure}{0.15\linewidth}
    \includegraphics[width=\textwidth]{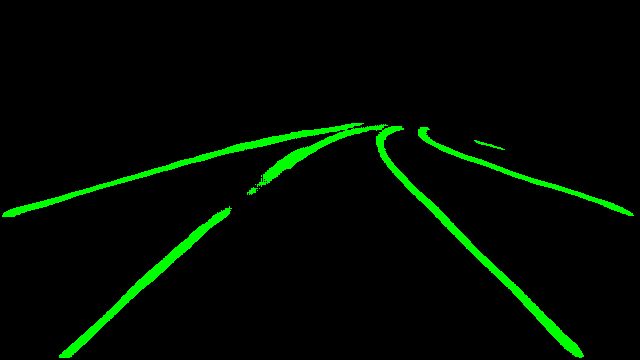}
  \end{subfigure}
  \hfill
  \begin{subfigure}{0.15\linewidth}
    \includegraphics[width=\textwidth]{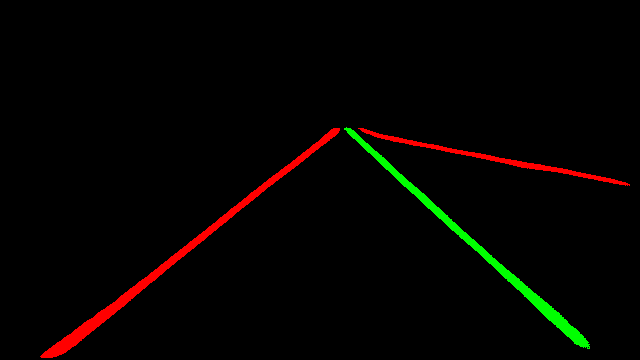}
  \end{subfigure}
  \hfill
  \begin{subfigure}{0.15\linewidth}
    \includegraphics[width=\textwidth]{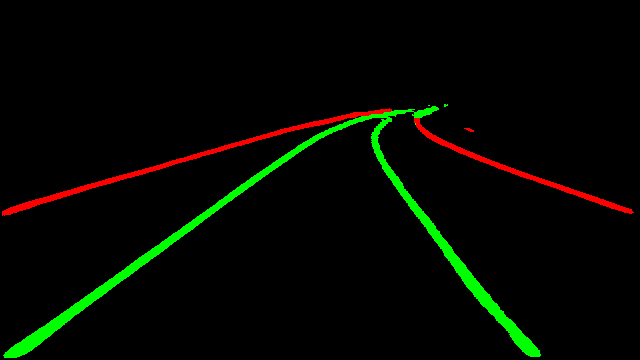}
  \end{subfigure}
  \hfill
\end{subfigure}
\caption{Qualitative results on the test set. From top to bottom: original image, instance segmentation, classification. For instance segmentation, different colors represent different boundaries. For classification, green represents dashed lanes, yellow double-dashed, red continuous.}
\label{results_images}

\end{figure}
\begin{table}
  \caption{\upshape{Ablation study on descriptor size.}}
  \label{table:results}
  \setlength{\tabcolsep}{4\tabcolsep}
  \centering

  \begin{tabular}{ *{4}{c} }
    \toprule
    \textbf{Descriptor size} & \textbf{Acc. (two classes)} &  \textbf{Acc. (three classes)}\\
    \midrule
    \(256\times256\) & \textbf{0.9698} & \textbf{0.9600} \\
    \(128\times128\) & 0.9596 & \textbf{0.9600} \\
    \(64\times64\) & 0.9519 & 0.9443 \\
    \(32\times32\) & 0.9527 & 0.9436 \\
    \(16\times16\) & 0.9359 & 0.9203 \\
    \bottomrule
  \end{tabular}
  \vspace{1pt}
\end{table}
 
As it is visible, it is possible to achieve better performances increasing the descriptor spatial resolution. However, this leads to a major occupation of GPU RAM. On the other hand, our results demonstrate that it is possible to achieve satifying accuracies with only 256 points. 
 
\section{Conclusions}
In this work, we presented a novel approach to lane boundary identification and classification in a end-to-end deep learning fashion. With our method, it is possible to achieve high accuracy in both tasks, in real-time. We formalized a descriptor extraction strategy that is useful when it is needed to combine instance segmentation and classification without relying on two-step detection networks. Furthermore, we performed an ablation study on the descriptor size, in order to define the tradeoff between detection accuracy and needed GPU RAM. 
 
\bibliographystyle{splncs}

\bibliography{bibliography}

\end{document}